# Dynamic Task Space Control Enables Soft Manipulators to Perform Real-World Tasks

*Oliver Fischer,\* Yasunori Toshimitsu, Amirhossein Kazemipour, and Robert K. Katzschmann\**

Dynamic motions are a key feature of robotic arms, enabling them to perform tasks quickly and efficiently. A majority of real-world soft robots rely on a quasistatic control approach, without taking dynamic characteristics into consideration. As a result, robots with this restriction move slowly and are not able to adequately deal with forces, such as handling unexpected perturbations or manipulating objects. A dynamic approach to control and modeling would allow soft robots to move faster and handle external forces more efficiently. In previous studies, different aspects of modeling, dynamic control, and physical implementation have been examined separately, while their combination has yet to be thoroughly investigated. Herein, the accuracy of the dynamic model of the multisegment continuum robot is improved by adding new elements that include variable stiffness and actuation behavior. Then, this improved model is integrated with state-of-the-art system identification and dynamic task space control and experiments are performed to validate this combination on a real-world manipulator. As a means of encouraging future research on dynamic control for soft robotic manipulators, the source code is made available for modeling, control, and system identification, along with the recipes for fabricating the manipulator.

## 1. Introduction

An emerging alternative to conventional rigid robotics is the field of soft robotics.[1–4] Soft robots adopt the principle of physical artificial intelligence to achieve an inherently compliant and embodied robotic behavior.[5,6] In soft robotics, robots are fabricated mainly from functional soft materials[7–10] to achieve inherently adaptive and intelligent characteristics. The inherent compliance of soft robots enables various behaviors that were previously difficult for conventional, rigid-linked robots to achieve, such as navigating through tight environments or gentle gripping.[11–13] Continuum robots[14–16] have shown improved adaptability and safety when moving from rigid to soft actuator materials and support structures. However, it is challenging to model continuum robots' behavior due to the infinite dimensionality of their state space.

There have been several approaches to construct a dynamic model for soft continuum manipulators, akin to the manipulator equation.[17,18] Godage et al.[19] derived the dynamic model of a continuum manipulator through integral Lagrangian formulation with the assumption of continuous mass distribution along the arm. However, the closed-form expressions of dynamic terms become complicated as the number of segments increases, making the model unpractical for using it in real-time control. In order to simplify the dynamic model, the mass distribution of each segment was approximated into a single lumped mass. Under this assumption, Falkenhahn et al. derived the dynamic model for a continuously bending manipulator (Festo Bionic Handling Assistant) where the lumped masses are assumed to be concentrated in the tips of the soft continuum sections.[20] With the addition of a dynamic model, they showed acceleration-level control methods in joint space. In a later work, Falkenhahn et al. explicitly considered valve dynamics to achieve higher model accuracy.[21] Curvature space methods can also be combined with inverse kinematic approaches to control real-world coordinates.[22,23] This combination was shown by Gong et al.,[24] who used an underwater continuum arm to grab samples. Alternatively, one can use machine learning methods to obtain a model of the dynamics. Thuruthel et al. use a recurrent neural network to approximate the dynamics, and learn a controller.[25] However, the blackbox nature of neural networks is undesirable for control.

Alternatively, it is possible to compute the dynamic parameters of a continuously bending soft body by using an augmented rigid body model to approximate its kinematic and dynamic characteristics.[26,27] This model supplements the piecewise constant curvature (PCC) model by adding a rigid link model and mass

O. Fischer, Y. Toshimitsu, A. Kazemipour, R. K. Katzschmann
Department of Mechanical and Process Engineering
Soft Robotics Laboratory
Institute of Robotics and Intelligent Systems
ETH Zurich, Zurich 8092, Switzerland
E-mail: olivefi@ethz.ch; rkk@ethz.ch

The ORCID identification number(s) for the author(s) of this article can be found under https://doi.org/10.1002/aisy.202200024.

points. Dynamic parameters can then be obtained using methods previously designed for rigid link models. This model has been implemented on a planar manipulator[26,28] and on a 3D manipulator controlling in joint space.[27] In Della Santina et al.,[29] a simulation work, it was shown that the actuation space must have at least the same dimensions as the operational space, and dynamic operational space control was demonstrated in a simulated soft continuum arm model. Kapadia et al. also show task space for planar continuum manipulators.[30–32]

Many of the previous works that have applied task space control to physical arms have employed a quasistatic assumption using a locally approximated Jacobian, which is derived either from a model[33,34] or learned,[35–38] to perform tip-follower actuation using local kinematics. These works have not considered dynamic parameters such as gravity and inertia. Instead, they use local kinematics at every step to incrementally move toward the target. This strategy does not allow for quick movements and skillful force application, as the actuation steps must be kept small to prevent the robot from oscillating.

Dynamic approaches to task space control of real-world robots have been investigated in Mustaza et al.,[39] which uses a dynamic model to control a single continuum segment in task space, and Franco et al.,[40] which investigates the step response of a two-segment continuum robot in the real world using a dynamic model. In our previous work,[41] we propose an adaptive sliding-mode controller which we validated on our soft manipulator. However, an accurate model which does not require adaption is still desirable.

In comparison, rigid manipulators are simpler in their structural composition and their rigid body dynamics can be accurately described.[42] Indeed, accurately identified models and extensive control theory allow rigid manipulators to quickly move through their task space.[43] Among these control schemes, operational space control formulates the dynamics in an operational space, and allows for accurate force application and dynamic motion while fulfilling the task given to the controller in the operational space.[44,45]

Experimental validation has been performed on real-world robots in Franco et al., who investigate the step response of a dynamically modeled soft manipulator.[40] Thuruthel et al. experimentally validate the dynamic response of a manipulator whose dynamics were modeled with a recurrent neural network.[25] Godage et al. validated their dynamic model of a soft manipulator in the real world, but did not use it for control.[46]

So far, previous studies have focused on individual aspects such as dynamic modeling, task space control, or physical validation of soft manipulators, with the exception of Franco et al., who investigated the task space step response. However, as a step toward unification, all of these aspects combined have yet to be investigated together for the task space dynamic response.

In this work, we combine dynamic modeling, dynamic task space control, and system identification methods to experimentally validate the integration of these elements on a real-world soft manipulator. To improve the accuracy of the dynamic model, we add new elements, i.e., the phase and magnitude adjustment elements, to the dynamic model. We then investigate their effect on task space control performance. We perform multiple experiments to investigate the performance of dynamic task space control on a real-world robot. We demonstrate multiple potential applications for soft manipulators such as pick-and-place, throwing objects, and drawing a straight line (**Figure 1**). We publish our modular codebase, design files, and fabrication recipes, enabling other researchers to validate our work on their manipulators and to improve over our methods. Our contributions can be summarized as follows: 1) We propose and validate new stiffness and actuation elements for the dynamic model of a soft manipulator. 2) We integrate our improved model with state-of-the-art system identification and dynamic control of soft continuum manipulators for real-world applications. 3) We perform a multitude of experiments to validate dynamic task space trajectory control on a real-world soft manipulator moving in 3D. 4) We compare dynamic and quasistatic control approaches for real-world continuum robots with respect to tracking performance. 5) We make available a modular codebase for dynamic control of real-world soft manipulators, as well as the design files and fabrication recipe.

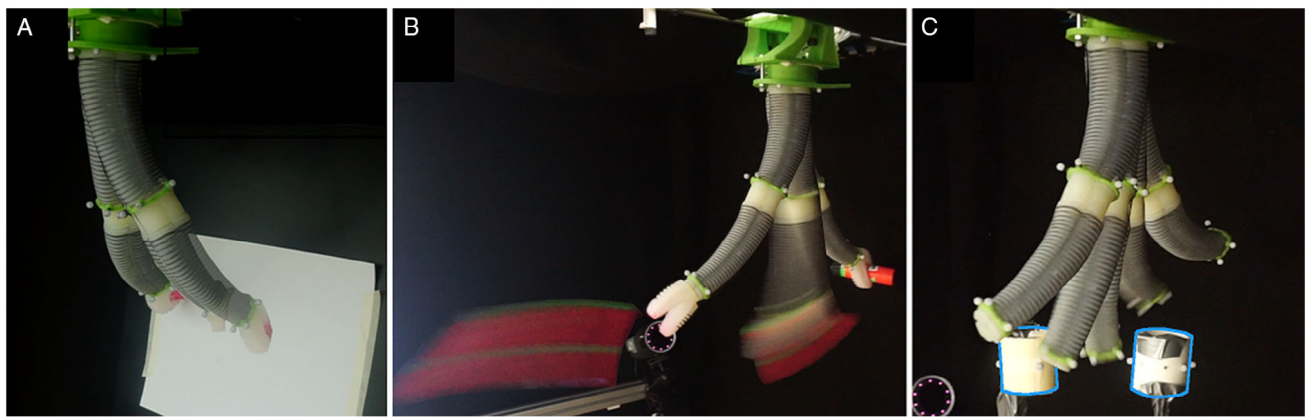

**Figure 1.** A soft manipulator performs various dynamic tasks. A) The manipulator draws with a grasped piece of chalk by moving from left to right. Motion snapshots are visually overlaid and spaced in time with 2 s intervals. B) The manipulator throws an object in a controlled manner. The motion from right and to left is depicted with snapshots at intervals of 130 ms. C) The manipulator avoids multiple obstacles while moving from right to left. Motion snapshots are spaced with 2 s intervals.

## 2. Experimental Section

### 2.1. System Architecture

The flow diagram of the full system is shown in **Figure 2**. The soft manipulator used in the experiments is the Soft Proprioceptive Arm (SoPrA) manipulator that was introduced by Toshimitsu et al.[47]. The SoPrA manipulator is cast in silicone, and it is fiber-reinforced to prevent bubbling and control the actuation behavior.[48] The arm's length is 0.27 m and it weighs 299 g in total. The gripper weighs 24 g and is included in the arm's total weight. The manipulator has two continuum segments. Each segment contains three actuation chambers. A soft silicone gripper is attached to the tip of the manipulator. Silicone tubing runs through the arm to actuate the chambers and gripper. Spherical reflective markers are mounted at the base of the arm and around the tip of each actuated segment. Task space objects are also marked with reflective markers. The motion capture system consists of eight infrared cameras, all of which are mounted around the manipulator and connected to a laptop that runs motion capture software (Miqus M3, Qualisys AB). The motion capture data are used to estimate the manipulator's curvature and velocity as well as the locations of the objects. The potential fields are derived from the objects' locations, and the resulting manipulator acceleration is computed. The controller then uses the location, velocity, and desired acceleration to calculate the desired pressure, which is then output to the valve system. The valve system is able to output up to 2 bars of pressure, but for this work we never exceeded 600 mbar to avoid ruptures in the manipulator (MBA-FB-VI, 0–2 bar range, 1% accuracy, Festo SE & Co. KG). The desired pressures are output to seven channels (three for each of the two segments and one for the gripper) and regulated with an internal proportional-integral-derivative controller in the valve system.

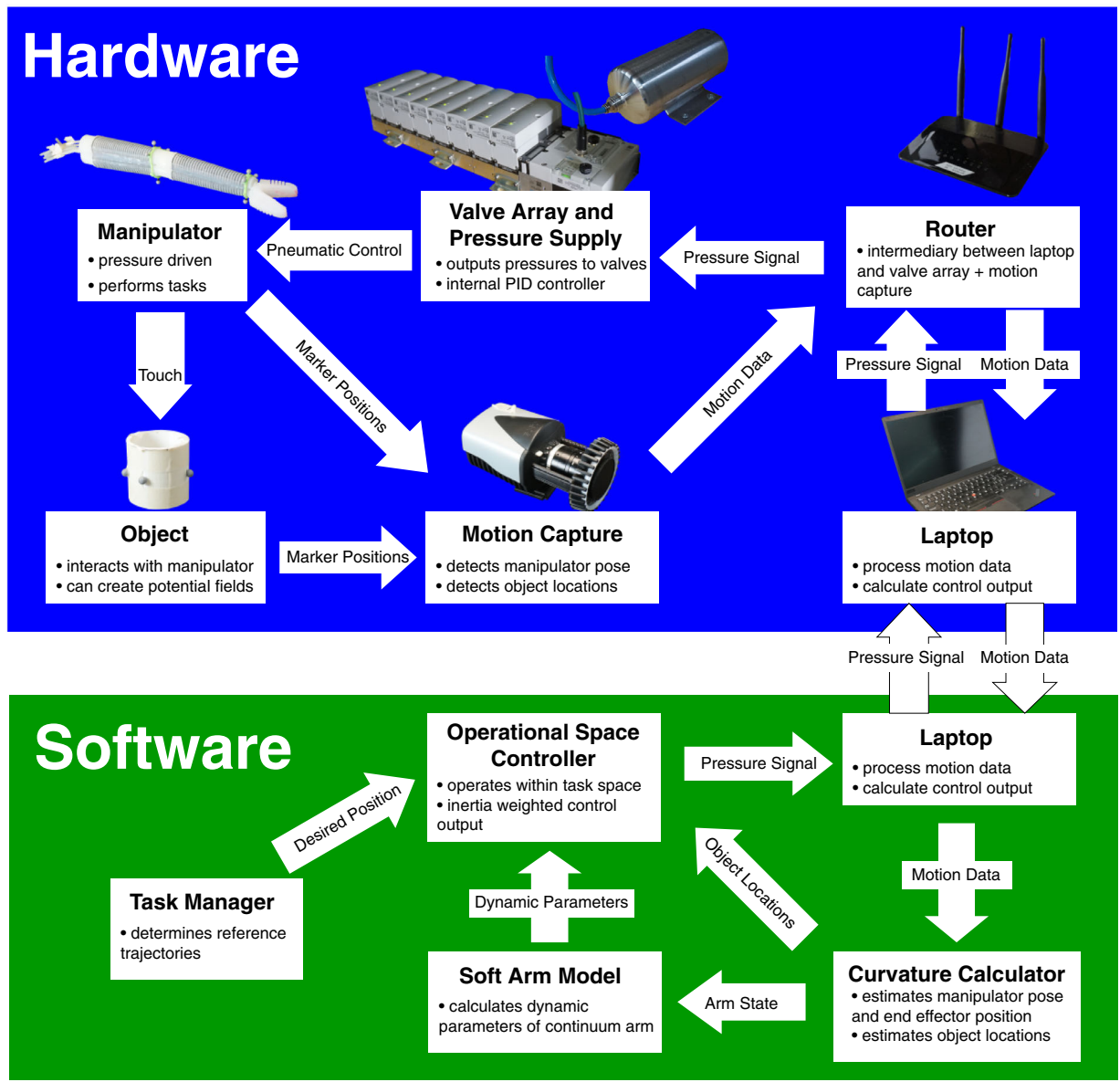

**Figure 2.** The signal flow of the complete system. The router acts as an intermediary between the controller on the laptop and the physical system. Motion capture detects the manipulator pose, which is used to determine dynamic parameters such as gravity and inertia. The pose and parameters are used with an operational space formulation to generate a pressure signal for the manipulator. The pressure signal is then sent to the valve array, which internally regulates pressure to the desired value.

### 2.2. Dynamic Model and its Adjustments

The manipulator is split into $N_{seg}$ controllable segments ($N_{seg} = 2$ for the experiments). According to the PCC approach,[49] the curvature of a single segment is constant. Traditionally, PCC segments are described by the variables $\phi$ and $\theta$, which describe the angle in plane of bending and curvature, respectively. We opt for the method used by Toshimitsu et al.,[47] which uses the following parameters: $\theta_x = \theta \cos(\phi)$ and $\theta_y = \theta \sin(\phi)$. These parameters describe the curvature in the $x$ and $y$ direction, respectively. They were chosen for easier understanding and identification of the dynamic equation matrices. Using our coordinates, we can describe the dynamic model of the soft manipulator

$$Ap_{xy} + J^T f_{ext} = B(q)\ddot{q} + c(q, \dot{q}) + g(q) + Kq + D\dot{q} \quad (1)$$

where $A$ maps the pressure $p_{xy}$ to generalized torques, $J$ is the Jacobian, and $f_{ext}$ is the force at the tip of the manipulator. $q$ is the curvature space parameterization of the manipulator and has a vector size of $2N_{seg}$. The corresponding values for each segment are $\theta_x$ and $\theta_y$. The dynamic parameters $B(q)$, $c(q, \dot{q})$, and $g(q)$ correspond to the inertia matrix, the coriolis/centripetal torque vector, and the gravity torque vector, respectively.

To calculate the dynamic parameters $B(q)$, $g(q)$, and $c(q, \dot{q})$ for the PCC model, we create an augmented rigid body model like the one that was originally proposed by Della Santina et al.[26]. We extended this model to the third dimension in a previous work.[27] This augmented rigid body model consists of five joints per PCC section. The configuration of the joints is computed from the PCC coordinates that are used in Toshimitsu et al.[47]. By representing each PCC element as a rigid body model, we are able to use the standard robotics library Drake[50] to compute the dynamic parameters of the rigid body model in rigid body coordinates $\xi$. These coordinates are then transformed back into configuration space of the PCC model $q$ by using the relation between $q$ and the augmented rigid body's joint angles $\xi$. This gives us the dynamic parameters in PCC coordinate parameterization, which can be used for control.

Based on a first-order approximation, we assumed that the bending stiffness of the soft manipulator $K$ was constant, regardless of bending state in the previously proposed model.[47]

However, the bending magnitude varied by as much as 33.6% depending on the angle of bending. This likely results from both the cross-sectional shape of the robot, which differs from that of the model due to manufacturing limitations, and the first-order approximation used in Toshimitsu et al.,[47] which does not hold up at higher degrees of bending. Additionally, fabrication errors and tapering cause the stiffness to vary between the chambers. This introduces a disparity between the model and reality: the torque inputs are of equal magnitude, but a different direction of actuation causes different magnitudes of deformation, despite the model predicting an equal deformation.

To compensate for this, we introduce the magnitude adjustment element into the stiffness matrix $K$. We define the magnitude adjustment element as a function of the angle of bending plane, $\phi$. To identify the variant parameter, we begin by assuming an invariant stiffness and identify this invariant term with a constant stiffness fit to quasistatic poses. Then, the arm is actuated with desired torques of equal magnitude for all angles in a rotating motion. Due to the aforementioned stiffness differences, the amount of bending deformation varies according to the angle of the bending plane $\phi$. The radial intensity can then be fitted to a function $f(\phi(q))$. It was heuristically determined that fitting a third degree polynomial to each 120° sweep covered by the three pneumatic chambers would provide a satisfactory characterization of the radial error. We premultiply the stiffness matrix $K$ with the inverse of the radial magnitude function $\frac{1}{f(\phi(q))}$ to obtain the angle-dependant stiffness matrix $K(\phi(q))$.

There is also an error in the desired plane of bending $\phi_{des}$ and in the one observed, $\phi_{meas}$. This error is corrected with a phase adjustment element. Again, we determine the error function, which can be expressed as $g(\phi_{des}) = \phi_{des} - \phi_{meas}$, by inputting feedforward pressures for $\phi \in [0, 360)$. Similar to the magnitude adjustment element, this error can be described with a third degree polynomial with regard to the angle $\phi$. We compensate for this error by adding an extra layer during actuation: by offsetting the input's direction with the angular error $\Delta\phi$, we can obtain an actuation signal that will produce the desired angle. We extract the intended direction of actuation from $p_{xy}$ and then rotate $p_{xy}$ by the angle corresponding to the error function $g(\phi)$. This way, the pressure's magnitude remains untouched. This transformation can be expressed as a rotational matrix $R(-g(\phi(p_{xy})))$, which we directly include in $A_{xy}$. This results in an actuation matrix that varies in behavior dependent on configuration: $A_{xy}(p_{xy}) = R(-g(\phi(p_{xy})))A_{xy}$.

### 2.3. System Identification

To operate a real-world manipulator robustly, the physical parameters must be identified. Simultaneous estimation of all parameters is theoretically possible, but risks fitting a wrong model. We therefore choose a hierarchical approach to parameter identification. The main parameters which must be identified for real-world are actuation, gravity, inertia, stiffness, and damping losses.

First, we measure length, diameters, and weight of each manipulator segment and connector piece. They are used by Drake[50] to calculate inertia, gravity, and coriolis parameters of the manipulator. These parameters are taken as ground truths due to the accuracy with which we can measure length and weight.

We first identify the stiffness parameter $K$, as it requires only a single value per segment. It is theoretically possible to derive the stiffness term $K$ from the shear modulus of the silicone. However in practice, manufacturing imperfections such as silicone bubbling and anisotropic settling cause the observed stiffness to slightly vary from the expected. We therefore identify the true stiffness under a quasistatic assumption: $\dot{q} = 0$, $\ddot{q} = 0$. This simplifies the dynamic equation to

$$Ap = g(q) + Kq \qquad (2)$$

where $g(q)$ is already known. To avoid simultaneous parameter identification, we use an estimate for $A$. We use design schematics, measured lengths, and measured diameters to estimate chambers' internal surface areas and distance from manipulator center. These are multiplied to obtain $A$. The manipulator then receives feedforward pressures to its chambers. We wait 10 s to ensure the manipulator is at rest and that the quasistatic assumption holds. Then, measurements are taken for pressures, manipulator curvatures, and corresponding gravity. This is repeated until a desired amount of measurements are reached. In this work, we used 40 poses per manipulator segment. We can then fit a "true" shear modulus using a least squares fit

$$K_{\text{true}} = q_{\text{meas}}^+ (Ap_{\text{meas}} - g_{\text{meas}}) \qquad (3)$$

where $^+$ denotes the Moore–Penrose pseudoinverse.

We now identify $A$, which is more complex, requiring 6 values per segment: each of the three chambers has an effect $x$- and $y$-directional curvature. We again make a quasistatic assumption and give the manipulator feedforward pressures, and then measure the pressures, curvatures, and corresponding gravity term. In this work, we used 360 poses per manipulator segment. We assume the previously calculated $K$ to be correct and fit $A$ using least-squares

$$A_{\text{true}} = p_{\text{meas}}^+ (g_{\text{meas}} + Kq_{\text{meas}}) \qquad (4)$$

We then reidentify $K$ and then $A$ using the obtain parameters to ensure they are correct. One can repeat the reidentification process infinitely to converge toward perfect values, but we found the values to change by less than 1% upon second identification and therefore did not further reidentify.

The magnitude adjustment element and phase adjustment element are then similarly identified by making a quasistatic assumption, giving feedforward pressures, and performing a least-squares fit.

Finally, we identify the damping parameter $D$ heuristically by controlling the manipulator in state-space using the model. The damping parameter is updated online while the control error is observed, and the value which produces the least error is chosen. We attempted to fit the value with a feedforward scheme similar to the previous parameter identifications as well as by releasing the manipulator and letting it swing, but were not successful using these approaches.

### 2.4. Controller Design

In this section, we detail how we have applied a force-based operational space controller to the soft robotic arm so that it can take advantage of its continuous body to fulfill tasks.

Our control law is a proportional-derivative controller in the form of

$$\ddot{x}_{\text{ref}} = k_p(x_{\text{des}} - x) + k_d(\dot{x}_{\text{des}} - \dot{x}) + \ddot{x}_{\text{des}} + \ddot{x}_{\text{pot}} \qquad (5)$$

where $x$ is the tip position in Cartesian coordinates, $x_{\text{des}}$ is the desired tip position, and $k_p$ and $k_d$ are gain terms for the position and velocity, respectively. $\ddot{x}_{\text{pot}}$ is the acceleration caused by potential fields. Additionally, the gains are made to saturate after reaching a set magnitude to prevent control inputs which could lead to unstable behavior.

The potential field acceleration used for avoiding obstacles, $\ddot{x}_{\text{pot}}$, is obtained as follows

$$\ddot{x}_{\text{pot}} = \begin{cases} k_{\text{pot}} \frac{1}{|\rho| - r_0} \frac{\rho}{|\rho|} & \text{if } |\rho| < r_m \\ 0 & \text{otherwise} \end{cases} \qquad (6)$$

where $k_{\text{pot}}$ is the gain of the potential field, $\rho$ is the Cartesian vector from the center of the potential field to the tip of the manipulator, $r_m$ is the cutoff radius of the potential field, after which it no longer affects the manipulator, and $r_0$ is the radius of the object that causes the potential field.

Objects are approximated as perfect spheres. As the tip of the manipulator approaches the object, the potential field acceleration $\ddot{x}_{\text{pot}}$, which is normal to the object, increases as the manipulator approaches the object. This leads to the tip of the manipulator being deflected away from the object. Similar to control gains, $\ddot{x}_{\text{pot}}$ saturates to prevent unstable behavior.

Having now obtained the reference acceleration $\ddot{x}_{\text{ref}}$, we perform a sequence of transformations: acceleration to a force in operational space, force to generalized torque, and then generalized torque to pressure. We thereby make use of an operational space formulation[42,43]

$$\begin{aligned} \tau_{\text{ref}} &= B(q)J^+(\ddot{x}_{\text{ref}} - \dot{J}\dot{q}) + g(q) + c(q,\dot{q}) + K(q)q + D\dot{q} \\ &\quad + (I - J^{+T}J^T)\tau_{\text{null}} \\ p &= A(q)^+(\tau_{\text{ref}} + f_{\text{supp}}) \end{aligned} \qquad (7)$$

where $f_{\text{supp}}$ is a vector that contains supplementary forces such as the desired tip force. $(I - J^{+T}J^T)\tau_{\text{null}}$ is the nullspace term, which utilizes redundant degrees of freedom to perform secondary tasks. In our work, we use the nullspace input $\tau_{\text{null}} = -\alpha q - \beta \dot{q}$, which drives the manipulator to the straightest possible configuration and dampens oscillations. $\alpha$ and $\beta$ are the straightness and damping gains, respectively.

## 3. Results

### 3.1. New Dynamic Model Elements

A real-world soft manipulator suffers from unmodeled fabrication imperfections which are not found in simulation. In particular, both stiffness and actuation can vary with configuration, which is not considered in previous simulation works. To consider these model parameters, we introduce a phase adjustment element and a magnitude adjustment element, which were explained in detail in Section 2.2. These elements serve the purpose of reducing directional errors resulting from model anisotropy. The effects of the adjustment elements are vital to understanding our modeling contribution and are therefore presented first.

We performed static feedforward experiments to characterize the robot's observed behavior and we compared it with the behavior of an unmodified model. We gave the manipulator pressure inputs of equal magnitude while varying the desired actuation direction from 0° to 360°

$$p_x = k \begin{pmatrix} \cos \phi \\ \sin \phi \end{pmatrix}, \text{ where } \phi \in \{0, 360\} \quad (8)$$

where $k$ was a constant with the value 500 mbar. In doing this, we identify the error between the desired polar angle $\phi_{\text{des}}$ and the measured polar angle $\phi_{\text{meas}}$. We then introduced the newly characterized phase adjustment element to the model and investigated the new angular error. The results are plotted in **Figure 3A**. The qualitative effect of our phase adjustment element is shown in Figure 3B, in which both manipulators are performing open-loop control. The mean absolute error between the desired and measured polar angle for the default model was $7 \pm 6°$, with a peak absolute error of $23 \pm 1°$. The model using our phase adjustment element showed a mean absolute error of $1 \pm 1°$ and a peak absolute error of $5 \pm 1°$. The error of the corrected model was largest in regions where the uncorrected model's error was changing quickly. This indicated localized concentrations of instability.

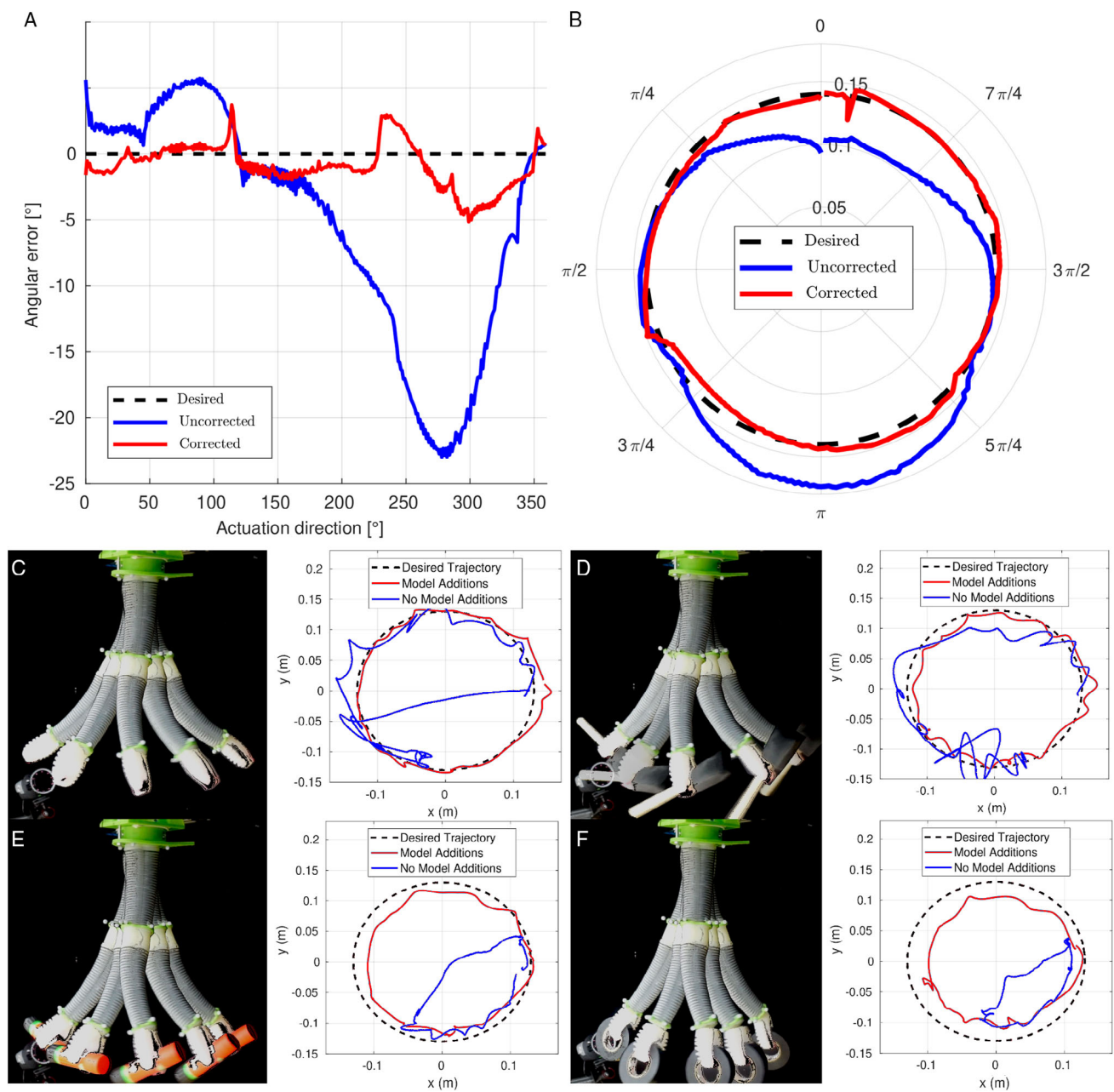

**Figure 3.** Model errors before and after new model elements were added. These errors directly impact absolute accuracy of control. A) The offset between the desired angle of actuation and the measured angle of actuation, with and without our new model elements. B) The offset between desired and measured radial magnitude, with and without our new model elements. C–F) Trajectories tracked with and without our model elements under different tip loads. The loads were (C): 0 g, (D): 12 g, (E): 24 g, (F): 40 g.

We additionally performed experiments to determine the effect of our magnitude adjustment element on the accuracy of the model. The manipulator was commanded to compensate for gravity and stiffness at a constant curvature intensity while the polar angle was varied from 0° to 360°

$$p_x = A + (g(q) + Kq), \text{ where } q \in \begin{pmatrix} k\cos\phi \\ k\sin\phi \end{pmatrix} \text{ and } \phi \in \{0, 360\} \quad (9)$$

We then compared the expected curvature intensity to the measured curvature intensity. The experiment was conducted with a static assumption to eliminate possible errors that could have occurred as a result of the inertia, damping, and coriolis effects. The experiment was performed once without any model additions and once with the magnitude adjustment element. The results are plotted in Figure 3C. The qualitative effect of our magnitude adjustment element is shown in Figure 3D, in which both manipulators are performing open-loop control. Without our magnitude adjustment element, the mean absolute error between the desired and measured radial magnitude was $1.8 \pm 1.3$ cm, and the peak absolute error was $4.7 \pm 0.1$ cm. When the magnitude adjustment element was used, the mean absolute error was $0.3 \pm 0.3$ cm, and the peak absolute error was $1.8 \pm 0.1$ cm.

We investigated the effect of our the new model elements by tracking circular trajectories both with and without our model elements. The reference trajectory for the circle was

$$x_{\text{des}} = \begin{pmatrix} 0.15 \cdot \cos(c \cdot t) \\ 0.15 \cdot \sin(c \cdot t) \\ -0.23 \end{pmatrix} \quad \dot{x}_{\text{des}} = \frac{\partial x_{\text{des}}}{\partial t} \quad \ddot{x}_{\text{des}} = \frac{\partial \dot{x}_{\text{des}}}{\partial t} \quad (10)$$

where $c$ is a constant with a value of $2 \cdot \pi/8$. We investigated trajectories for different loads. The results are shown in Figure 3C–F, corresponding to loads of 0, 12, 24, and 40 g respectively. The model using our additional elements displays less tracking error and oscillation. At higher loads, the disparity between the models is greater.

### 3.2. Pick-and-Place

We examined a core behavior of robotic manipulators in the form of a pick-and-place task. The task was performed multiple times to examine repeatability. The manipulator was given a trajectory to a predefined point, at which a grape was located. The manipulator was then commanded to grab the grape, compensate for its weight, and given a second trajectory to a drop-off point over a small container. Finally, the manipulator was made to drop the

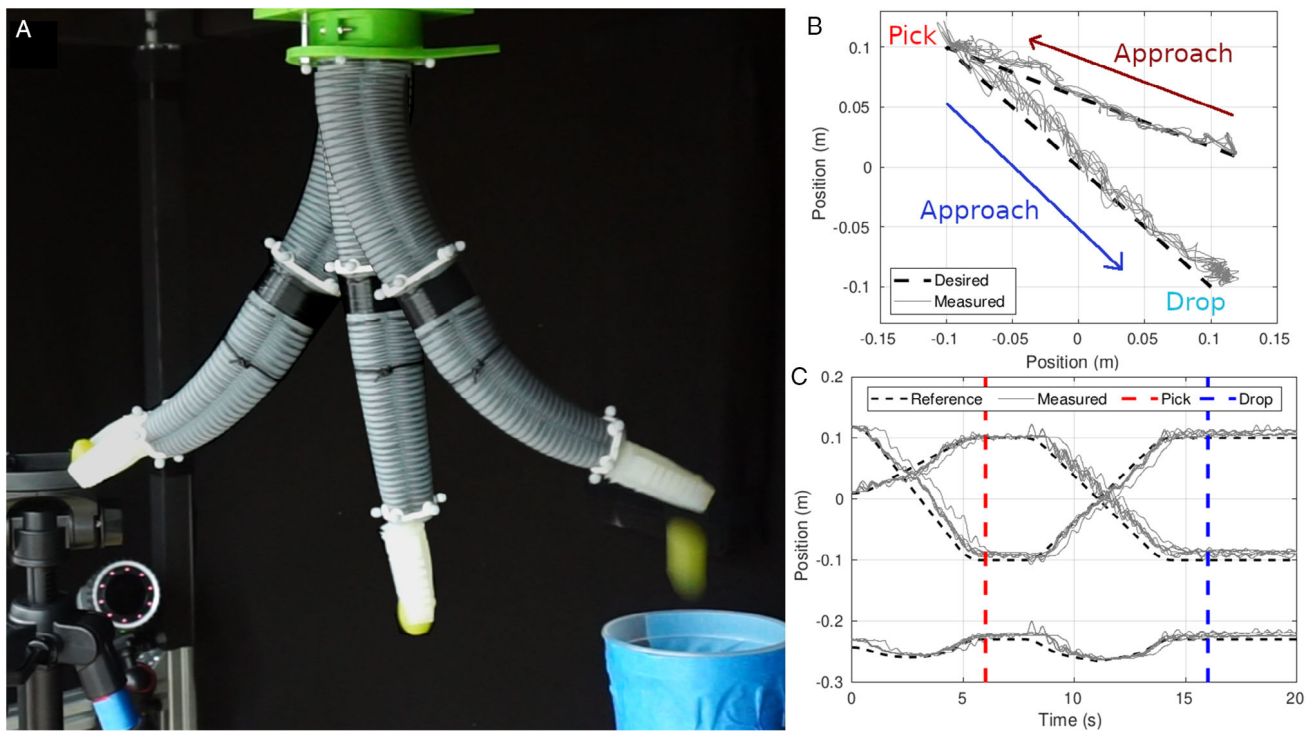

**Figure 4.** Soft manipulator picking up and dropping a grape in a controlled manner. The grape's position was assumed to be known. A) Manipulator traversing the workspace from picking position to dropping position, and dropping the grape in a cup. Motion frames are overlaid with 2.5 s intervals. B) Manipulator trajectories during the task. C) Manipulator coordinates plotted against their references.

grape into the container. The reference for the point-point movements was as follows

$$x_{des} = \frac{x_2 - x_1}{c} t \quad \dot{x}_{des} = \frac{x_2 - x_1}{c} \quad \ddot{x}_{des} = 0 \tag{11}$$

where $x_1$ is the starting point, $x_2$ is the end point, and $c$ is a constant describing desired time between the two points. The trajectories are plotted in **Figure 4**B,C. Only one pick point was investigated, as the restrictive nature of the workspace meant few locations were available for picking. Average tip error during the trials was $1.8\,cm \pm 0.6\,cm$. This is equivalent to $6.7\% \pm 2.2\%$ of total arm length. The task was performed successfully in six out of seven trials, with one failure where the gripper lost grip of the grape while moving toward the container. The trajectory was tracked successfully in all trials. The trajectory moves $\approx 50\,cm$ in $15\,s$, which is not possible with a quasistatic controller which achieves a maximum speed of $1.2\,cm\,s^{-1}$ when implemented on our manipulator.

### 3.3. Throwing

The manipulator was commanded to follow a linear trajectory with a high desired speed ($0.5\,m\,s^{-1}$) and to release a gripped object along the way, leading to a throwing behavior. We used Equation (11) for the trajectory input. Different objects with varying weights were thrown. We performed the experiment twice for each object: once with our model additions (magnitude adjustment element and phase adjustment element) and once without these elements, which is the current state of the art.[29] The results are plotted in **Figure 5**B–G). Blue lines indicate trajectories taken when using our new model elements, and yellow lines indicate trajectories taken without our new model elements. The objects thrown were: (B,C) tape roll weighing $11\,g$, (D,E) gluestick weighing $24\,g$, and (F,G) electric tape roll weighing $40\,g$.

When our additional model elements were included, the manipulator threw the objects and stabilized after the throwing motion. When using the model without our additional elements, the manipulator was still able to throw objects, but it was unable to stabilize after reaching the goal position due to a mismatch between the calculated dynamic parameters and the real-world state. The behavior appears not to change with different weights. Throwing is not possible with a quasistatic controller since such controllers are not able to reach high speeds: our manipulator achieved a max speed of $0.012\,\mathrm{m\,s^{-1}}$ when using a quasistatic controller.

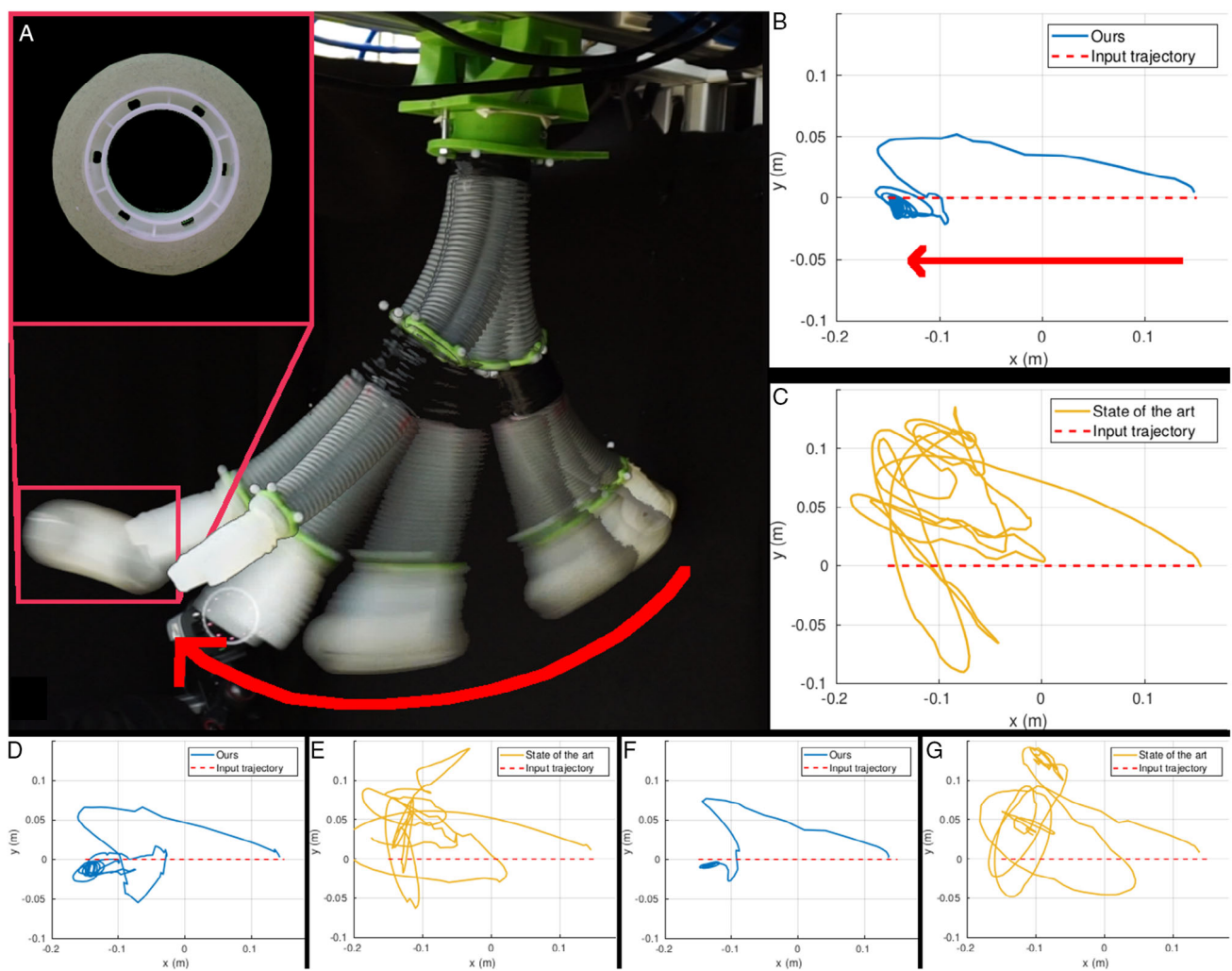

**Figure 5.** The soft manipulator's stabilization behavior while throwing various objects. All graphs displayed the first 5 s of manipulator behavior upon starting the throw. A) Manipulator throwing tape, shown in the top left corner. The images are spaced in 65 ms intervals. B,C) Trajectories taken by the manipulator when throwing tape weighing 11 g with and without our new model elements, respectively. D,E) Trajectories taken when throwing a gluestick weighing 24 g. F,G) Trajectories taken when throwing electric tape weighing 40 g.

### 3.4. Drawing

The manipulator was commanded to a position close to a blackboard setup, which was placed inside the task space. A soft gripper was mounted to the manipulator. The gripper grasped a piece of pink chalk, which is shown in **Figure 6**B. When the manipulator tip reached the blackboard's surface, we command a tip force of 1 N in the direction of the blackboard. Then, using the controller, we made the manipulator move in a straight line along the blackboard using the trajectory described in Equation (11) while applying a force in order to draw a line. The end effector followed the trajectory of a line, as shown in Figure 6C. The manipulator was able to track the line smoothly. The mean absolute vertical error was 0.34 cm. The error appeared mostly beneath the desired trajectory, which may result from the shape of the arm's task space. Results demonstrating repeatability of the task can be found in the Supporting Information.

### 3.5. Circular Tracking

A robot's ability to follow trajectories is essential in robotic manipulation. In this experiment, we compared our dynamic controller and a quasistatic controller with regard to trajectory tracking ability. We performed this experiment in both a vertical "stalactite" configuration and a horizontal "beam" configuration to investigate the manipulator's ability to deal with different gravitational effects. We also investigated different trajectory periods. The controllers tracked a low period trajectory, and a longer period trajectory, which was chosen at a speed which the quasistatic controller can follow. The circular trajectories were given as in Equation (10).

The quasistatic controller uses local kinematics to add a pressure differential to the current pressure

$$p_{x+1} = p_x + \alpha J^+(x_{\mathrm{des}} - x) \quad (12)$$

where $\alpha$ is a scaling factor. We applied a saturation function to the error $x_{\mathrm{des}} - x$ and chose $\alpha$ heuristically such that the manipulator could move at the quickest possible speed without oscillating out of control.

Experiments in horizontal configuration are shown in **Figure 7**A. When tracking a circle with an 8 s period, the quasistatic controller was unable to respond quickly enough to the trajectory. When tracking a circle with a 45 s period, the quasistatic controller showed reduced error compared with the dynamic controller. Both controllers displayed largest error in the z-direction, suggesting that gravity presented the largest issue for both controllers.

Experiments in vertical configuration are shown in Figure 7B. Again, the quasistatic controller was unable to track a trajectory with a low period. When tracking a trajectory with a 115 s period, both controllers displayed good tracking, with the quasistatic controller achieving slightly lower average error.

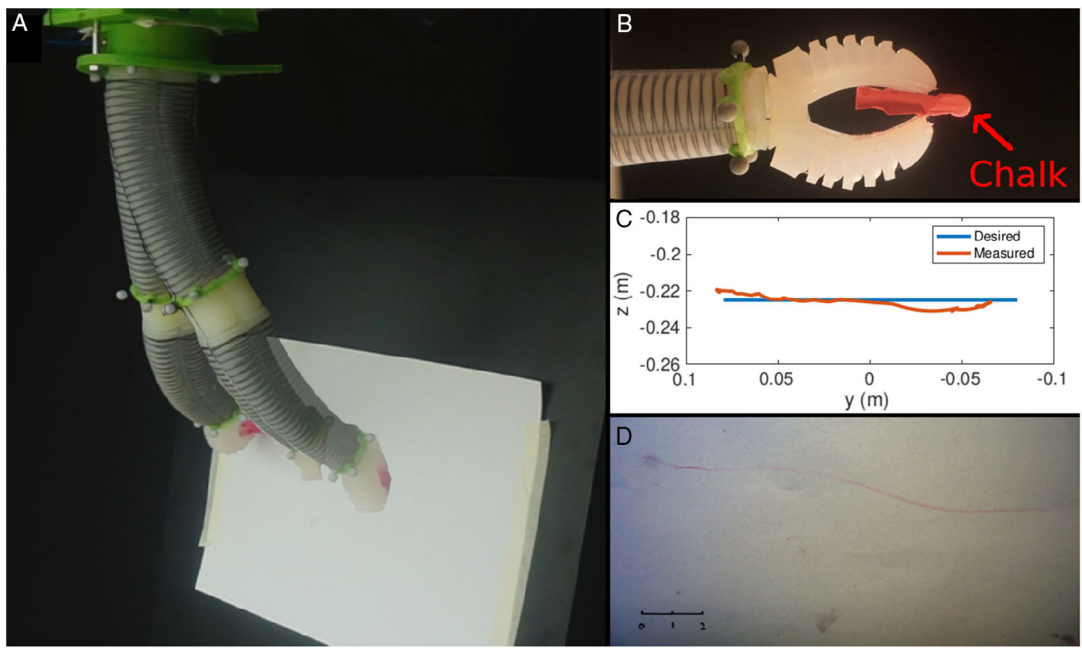

**Figure 6.** Soft manipulator demonstrating force application in a drawing task. A) Soft manipulator drawing a line with a piece of chalk. Images are spaced in time with 1.5 s intervals. B) Soft manipulator gripping a pink piece of chalk. C) Tracked trajectory, with desired trajectory referenced. D) Image of the drawn line.

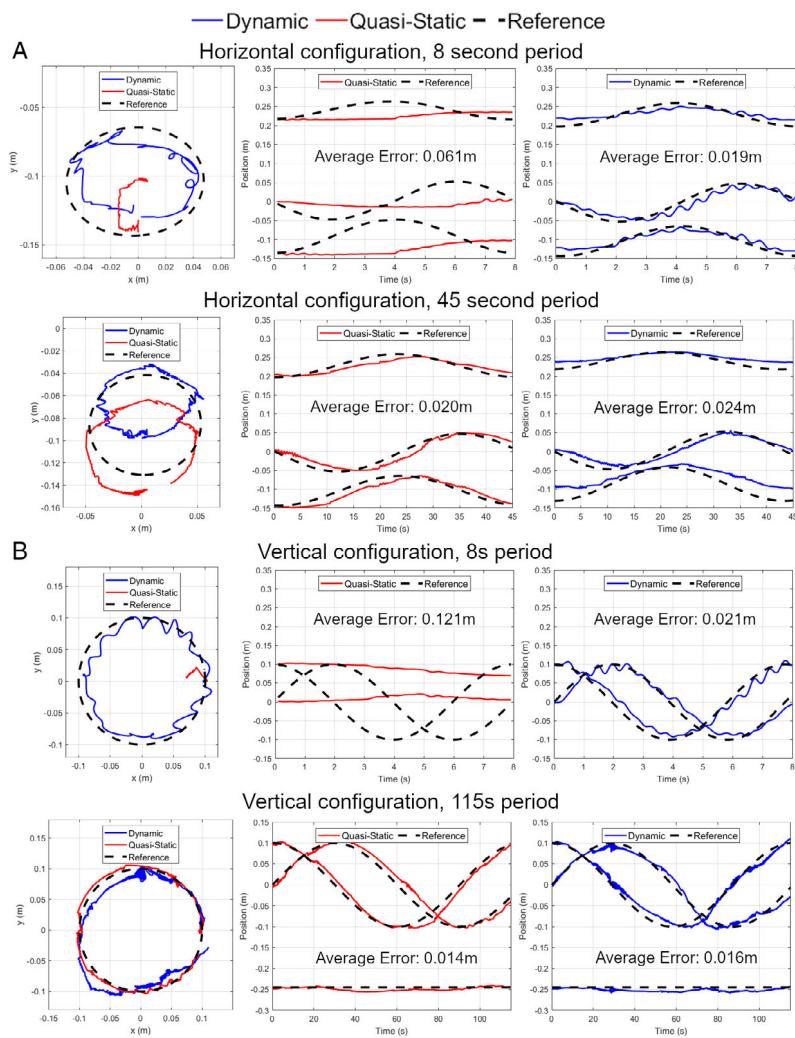
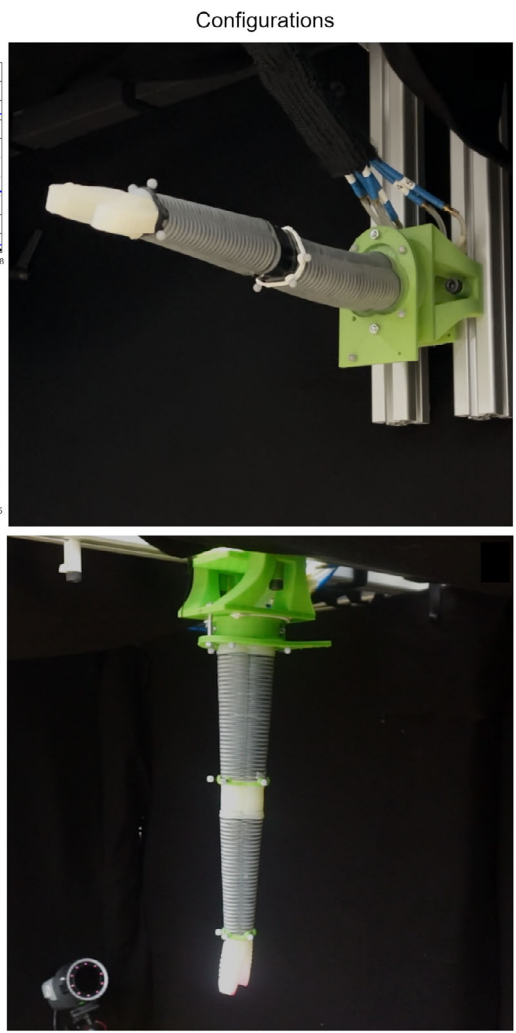

**Figure 7.** Circular tracking of the manipulator in different configurations. We varied both manipulator configuration and trajectory period to investigate different working conditions. A) Experiments with different periods in horizontal configuration. The manipulator is shown compensating for large gravity forces. B) Experiments with different periods in vertical configuration. The manipulator is shown hanging freely.

### 3.6. Disturbance and Step Response

We made the manipulator grip an object weighing 38 g and considered this as a tip force of 0.37 N in negative $z$-direction. The manipulator's responses to a the load while tracking are shown in Figure 8A. We recall that the tracking errors for the trajectories when not loaded were 1.4 cm for the quasistatic controller, and 1.6 cm for the dynamic controller. When loaded, the errors were 2.7 cm for the quasistatic controller, and 1.6 cm for the dynamic controller. The dynamic controller's error therefore did not increase upon adding a load, while the quasistatic controller's error nearly doubled.

We examined the step responses, as shown in Figure 8B. The dynamic controller reduced error a magnitude faster than the quasistatic controller; however, it displayed more oscillation. Furthermore, the quasistatic controller's steady-state error was slightly lower than the dynamic controller's.

We also investigated an outside disturbance, which was induced by hitting the manipulator with a stick. The responses are plotted in Figure 8C. The dynamic controller displayed less initial deviation and reduced error faster than the quasistatic controller. Both controllers oscillated heavily in response to the disturbance. Again, the quasistatic controller displays a slightly lower steady state error.

## 4. Discussion

### 4.1. Conclusion

The results show that the chosen approach advances the capabilities of soft manipulators. To the best of our knowledge, we were the first to dynamically control a soft manipulator's motion in task space in 3D and were able to use the dynamic model to apply tip forces in a drawing experiment. The addition of our new model elements, adjusting the actuation and stiffness to fit the nonlinear characteristics of the soft arm, improved model accuracy and tracking performance. We compared a dynamic control approach to a quasistatic approach. The quasistatic controller performed slightly better than the dynamic controller when tracking trajectories with long periods, but it was unable to follow faster trajectories. The dynamic controller was able to follow fast trajectories and showed no change in error when dealing with a tip load, whereas the quasistatic controller displayed nearly double the error. An analysis of step response and disturbance response reveals that the dynamic controller is able to respond approximately one order of magnitude faster than the quasistatic controller, while the errors remain comparable. We note that dynamic control's tip velocity is only limited by the capture speed of the motion tracking system, the accuracy of the PCC-based model, and the computation time, while quasistatic controllers are inherently limited in speed.

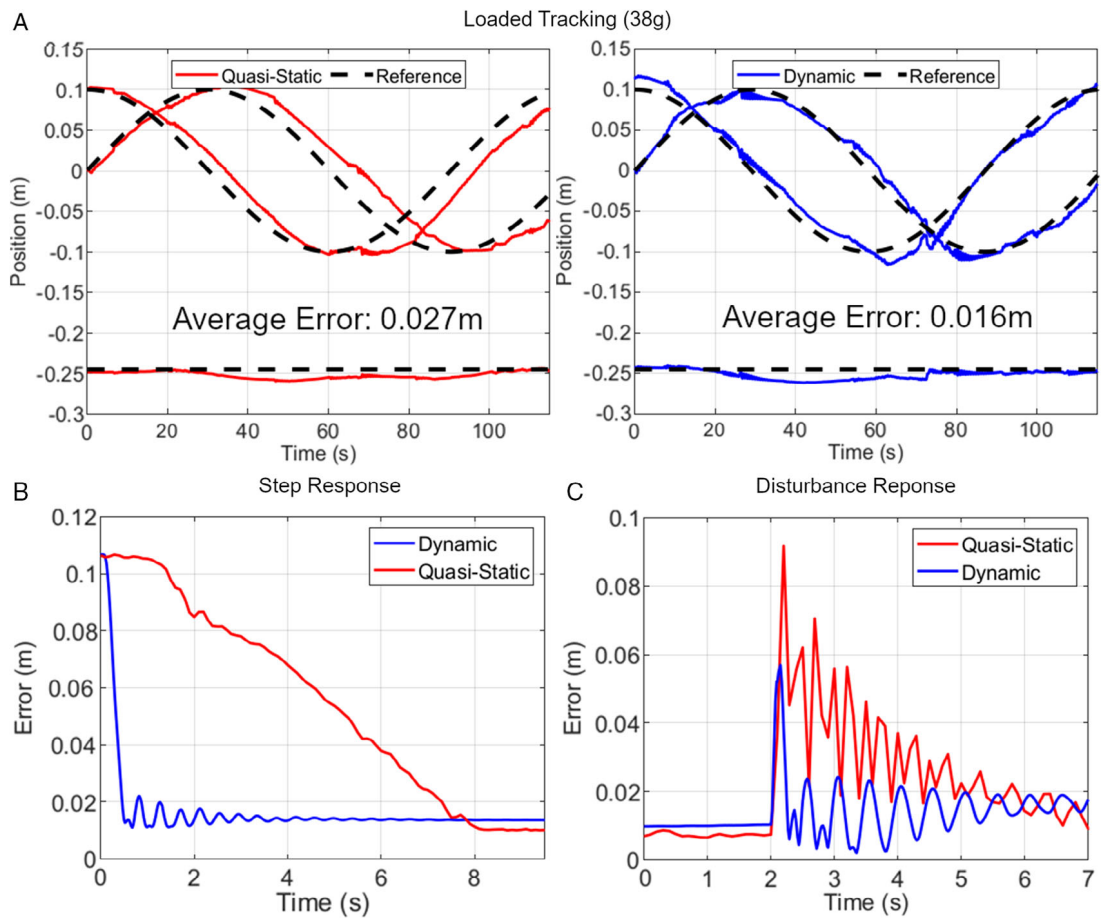

**Figure 8.** Manipulators operating in the real world may encounter situations in which they must move point-to-point, deal with an outside force, or operate under loaded conditions. In this experiment, we examined the different control approaches' behavior in these situations.

The throwing experiment clearly demonstrates the usefulness of our new elements for the dynamic model to perform velocity control. Without our new elements, the manipulator was unable to stabilize after reaching a high speed. When using our new model elements, the manipulator stabilized after throwing multiple objects of different weights. Our demonstration of a pick-and-place task summarizes our contributions: the manipulator moves at a higher velocity than possible with quasistatic control, and handles a load while moving toward the drop-off spot.

### 4.2. Limitations and Future Work

While we have demonstrated that dynamic control for soft manipulators is useful, the methods can still be shown in further scenarios. A limiting factor in our experimental investigation was the workspace of our continuum manipulator, a plot of which can be found in the Supporting Information. The workspace currently has a hemispherical shape, and limits the robot to a spherical plane ($r = $ const). It is essential to future investigations of soft manipulator control that this workspace is extended. An easy way to increase the workspace could be the addition of a prismatic joint at the base of the manipulator. This would allow the hemisphere to shift up and down, greatly increasing the positions the manipulator tip could reach. Online programming of the actuation behavior[51] could also help increase the workspace, as more shapes could be taken by the continuum segments. This would manifest itself as dynamically changing $A(q)$ matrix.

The piecewise constant curvature modeling approach also has limitations. While it acts as a suitable approximation of the robot's curvature properties, the actual robot does not display an ideal constant curvature bending. Instead, the robot bends with variable curvature. Additionally, an assumption is made that neither actuation nor stiffness varies with curvature intensity, which is also an approximation. The real robot is nonlinear with regard to curvature as a result of variation of cross section during actuation. While this inaccurate model is compensated through our closed-loop control approach, it nevertheless affects the tracking speed and accuracy, which can be seen in the dynamic controller's oscillations.

Alternatives to the PCC modeling for soft robotics have already been investigated with new methods such as Cosserat rods.[52–56] However, nonconstant curvature approaches with Cosserat rods are difficult to model for control, as systems must have finite dimensionality, and any Cosserat rod will be inherently underactuated. In future works, improved kinematic modeling approaches could be used in conjunction with a dynamic model (obtained from either an augmented rigid arm or a Lagrangian) to combine the dynamic properties of our arm with better kinematic modeling approaches. The nonlinear actuation and stiffness behavior could be modeled to improve model accuracy beyond what is currently possible. This could lead to an arm that is possible to control even in a model predictive control framework.

Our model correction approach could also be further improved. We currently add an experimentally determined error correction term to our actuation and stiffness matrices. These error correction terms smooth unknown modeling errors and require system identification of every manipulator we wish to control. In future works, we will eliminate further model error sources. Potential error sources include the anisotropic stiffness, differences in the wall thickness, and the manufacturing quality of the cast silicone. Moreover, supplementing or replacing the model with a finite element model of the manipulator could lead to a better state estimation, as the elastic body would be more precisely modeled.

Finally, we rely heavily on a motion capture system to gather both the robot configuration and the location of objects within task space. This choice is limiting due to the large amount of setup and cost required for a motion capture system. It also makes the robot impractical to be used outside of a laboratory setting. The dependency on motion capture systems may be addressed using cheaper sensor options such as embedded bend sensors[47] or inertial measurement units[57] to estimate the state of the manipulator. Objects within the task space could be found using a comparatively cheap depth sensing camera and previously developed object detection methods[58] to enable an integrated system without dependence on bulky and expensive motion capture setups.

## Supporting Information

Supporting Information is available from the Wiley Online Library or from the author.

## Acknowledgements

The authors are grateful for the help by Antares Yu Zhang and Ki Wan Wong in fabricating some of the manipulators. Further thanks go to Mike Yan Michelis, Barnabas Gavin Cangan, Thomas Buchner, Minghao Han, and Oncay Yasa for their helpful comments during the writing process.

## Conflict of Interest

The authors declare no conflict of interest.

## Data Availability Statement

The data that support the findings of this study are available from the corresponding author upon reasonable request.

## Keywords

dynamic control, manipulation, soft robots, system modeling

Received: January 25, 2022
Revised: July 25, 2022
Published online: